\documentclass[10pt,twocolumn,letterpaper]{article}

\usepackage{cvpr}              

\usepackage{amsmath,amsfonts,bm}

\def\eqref#1{equation~\ref{#1}}

\def\1{\bm{1}}

\DeclareMathAlphabet{\mathsfit}{\encodingdefault}{\sfdefault}{m}{sl}
\SetMathAlphabet{\mathsfit}{bold}{\encodingdefault}{\sfdefault}{bx}{n}

\def\gO{{\mathcal{O}}}

\def\gS{{\mathcal{S}}}

\def\gY{{\mathcal{Y}}}

\usepackage{url}

\usepackage{booktabs}       
\usepackage{amsfonts}       
\usepackage{nicefrac}       
\usepackage{microtype}      
\usepackage{xcolor}         
\usepackage{graphicx}
\usepackage{amsmath}
\usepackage{subcaption,booktabs}
\usepackage{array}
\usepackage{multirow}
\usepackage{listings}
\usepackage{sidecap}
\usepackage{amssymb}
\usepackage{mathtools}
\usepackage{csquotes}
\usepackage{multirow}
\usepackage{pifont}
\usepackage{wrapfig,lipsum}
\usepackage{makecell}
\usepackage{enumitem}

\usepackage{listings}
\definecolor{codegreen}{rgb}{0,0.6,0}
\definecolor{codegray}{rgb}{0.5,0.5,0.5}
\definecolor{codepurple}{rgb}{0.58,0,0.82}
\definecolor{backcolour}{rgb}{0.95,0.95,0.92}
\lstdefinestyle{mystyle}{
    backgroundcolor=\color{backcolour},   
    commentstyle=\color{codegreen},
    keywordstyle=\color{magenta},
    numberstyle=\tiny\color{codegray},
    stringstyle=\color{codepurple},
    basicstyle=\ttfamily\footnotesize,
    breakatwhitespace=false,         
    breaklines=true,                 
    captionpos=b,                    
    keepspaces=true,                 
    numbers=left,                    
    numbersep=5pt,                  
    showspaces=false,                
    showstringspaces=false,
    showtabs=false,                  
    tabsize=2
}
\newcommand{\cmark}{\color{teal}\ding{52}}
\newcommand{\xmark}{\color{red}\ding{56}}
\newcommand*{\ditto}{---\texttt{"}---}

\newcommand{\myparagraph}[1]{\noindent{\bf{#1}}}

\newcommand{\delete}[1]{\textcolor{cyan}{}}
\newcommand{\add}[1]{#1}

\newcommand{\ours}{FLM}
\newcommand{\oursFull}{Feasibility with Language Model}

\definecolor{cvprblue}{rgb}{0.21,0.49,0.74}
\usepackage[pagebackref,breaklinks,colorlinks,citecolor=cvprblue]{hyperref}

\def\paperID{31} 
\def\confName{ECCVW - OODCV}
\def\confYear{2024}

\title{\large Feasibility with Language Models for Open-World Compositional Zero-Shot Learning}

\author{
Jae Myung Kim$^{1,2,4}$
\;\;\;\; Stephan Alaniz$^{2,3,4}$
\;\;\;\; Cordelia Schmid$^{5}$
\;\;\;\; Zeynep Akata$^{2,3,4}$ 
\\
\\
\small{$^{1}$University of Tübingen}
\;\;\;\; \small{$^{2}$Helmholtz Munich}
\;\;\;\; \small{$^{3}$Technical University of Munich}  \\
\small{$^{4}$Munich Center for Machine Learning}
\;\;\;\; \small{$^{5}$Inria, Ecole normale sup\'erieure, CNRS, PSL Research University}
\vspace{-10pt}
}

\begin{document}
\maketitle

\begin{abstract}
Humans can easily tell if an attribute (also called state) is realistic, i.e., feasible, for an object, e.g. fire can be \textit{hot}, but it cannot be \textit{wet}.
In Open-World Compositional Zero-Shot Learning, when all possible state-object combinations are considered as unseen classes, zero-shot predictors tend to perform poorly.
Our work focuses on using external auxiliary knowledge to determine the feasibility of state-object combinations.
Our \oursFull\ (\ours) is a simple and effective approach that leverages Large Language Models (LLMs) to better comprehend the semantic relationships between states and objects. \ours\ involves querying an LLM about the feasibility of a given pair and retrieving the output logit for the positive answer. To mitigate potential misguidance of the LLM given that many of the state-object compositions are rare or completely infeasible, we observe that the in-context learning ability of LLMs is essential.
We present an extensive study identifying Vicuna and ChatGPT as best performing, and we demonstrate that our \ours\ consistently improves OW-CZSL performance across all three benchmarks.
\end{abstract}

\section{Introduction}
\label{sec:intro}

Humans have the ability to discern the feasibility of state-object pairs, effortlessly distinguishing between realistic and implausible combinations. For instance, while it is convincing for a \textit{fire} to be \textit{hot}, the notion of a \textit{wet fire} is nonsensical. Open-world compositional zero-shot learning (OW-CZSL) \citep{owczsl} seeks to emulate human-like understanding for compositional concepts. The task is to classify images to the correct state-object pair in the absence of explicit knowledge regarding the feasibility of the pairs in the candidate classes 
when the model is trained with a small subset of feasible pairs.
Prior works~\citep{owczsl, kgsp} proposed to remove possibly infeasible pairs from the label space using word vectors such as GloVe~\citep{glove} or using external resources such as ConceptNet~\citep{conceptnet}. While these approaches represent a step forward, open-world compositional zero-shot learning remains extremely challenging as these approaches are limited in their capability to capture the semantic relationships underlying many rare concept compositions. Our goal is to propose a more effective approach for determining the feasibility of state-object pairs even if they are rare.

Large language models (LLMs) recently demonstrated strong language comprehension capabilities across various NLP tasks~\citep{zhao2023survey}. In this work, we propose \oursFull\ (\ours) to predict the feasibility score of any state-object pair.
Concretely, we ask an LLM to give a binary response, i.e., "Yes" or "No", indicating the feasibility of the given state-object pair. The output logit for the positive answer would then be considered the feasibility score for the corresponding pair.

Inevitably, one challenge in using LLMs for feasibility prediction is that, provided without context a query could lead to many false negatives. Consider the ``dark fire'' class of the MIT-States~\citep{mitstates} dataset, that is considered feasible. Asking a LLM, whether ``dark fire'' exists, yields the answer ``No'', presumably because the state ``dark'' is not typically associated with bright objects. However, ``dark fire'' is a reasonable class in MIT-States, as humans assigned this label to highlight the dark surroundings and dim visual theme for these images of fire. To teach the LLM about the relevant context for image classification, we can inform the LLM of semantically similar and feasible compositions from the training set, such as ``dark lightning''. As a result, the LLM can correctly infer \textit{in-context} that the state "dark" can also be associated with ``fire''.

Our contributions are: 1) in \oursFull\ (\ours) we propose to leverage LLMs to predict the feasibility of state-object pairs in open-world CZSL with in-context guided prompts, 2) the  feasibility judgement via \ours\ better aligns with human-annotated ground truth, and can be integrated into any existing VLM, 3) \ours\ consistently improves CZSL performance over previous state-of-the-art methods on all three challenging benchmark datasets.

\begin{figure*}[t]
    \centering
    \includegraphics[width=0.9\linewidth]{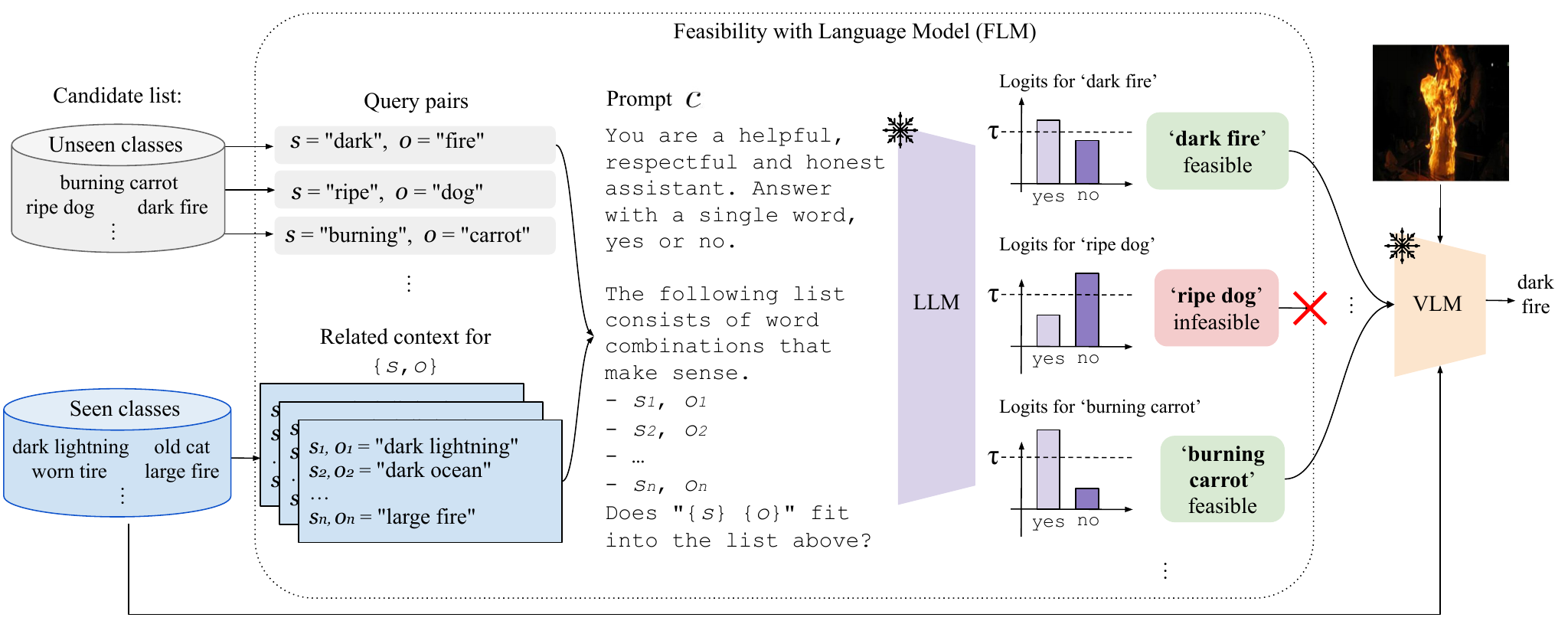} 
    \vspace{-2mm}
    \caption{The pipeline of our \oursFull\ (\ours) method. We constrict a prompt containing a list of related seen classes from the training set and a query to classify an unssen state-object pair as feasible. By comparing the LLM logit for the token ``Yes'' with a threshold $\tau$ we determine whether a pair is feasible, in which case it is used for OW-CZSL classification. 
    }
    \label{fig:main}
    \vspace{-2mm}
\end{figure*}

\section{Related work}

\add{
\myparagraph{CZSL.} CZSL aims to classify instances of state-object compositions not seen during training. Early approaches tackle CZSL by employing two separate classifiers for state and object primitives \citep{misra2017red,purushwalkam2019task}, treating the state as a transformation operator that modifies attributes of the object \citep{li2020symmetry,nagarajan2018attributes}, or leveraging Graph Convolutional Neural Networks to capture the relevance between state-object pairs \citep{cgqa,ruis2021independent}. Recently, VLMs like CLIP~\citep{clip} have been employed to address CZSL~\citep{coop, csp}.
}

\myparagraph{OW-CZSL.} Initially, CZSL models were evaluated exclusively on unseen classes. Then \cite{gzsl} introduced a generalized setting that considers both seen and unseen classes as potential labels, which was first used in CZSL by \cite{purushwalkam2019task}. The open-world setting~\citep{owczsl}, extends the output space to include all possible combinations of states and objects, which is a more challenging task due to the substantial increase in the number of potential candidates. Consequently, identifying and disregarding infeasible state-object pairs becomes crucial. Prior approaches have employed word vectors, i.e., CompCos~\citep{owczsl}, or external knowledge, i.e., KGSP~\citep{kgsp}, to determine the feasibility of pairs. In contrast, we propose to leverage state-of-the-art large language models to capture the feasibility of pairs more effectively.

\section{In-context Feasibility Prediction Framework}

In this section, we first describe the general setting of compositional zero-shot learning in \S\ref{subsec:preliminaries}. Next, we explain the novel utilization of large language models for predicting feasibility scores in \S\ref{subsec:method}.

\subsection{Open-World Compositional Zero-Shot Learning (OW-CZSL)}
\label{subsec:preliminaries}

CZSL aims to classify an image, where each class is a state-object combination. Given a set of states $\gS$ and objects $\gO$, the label space of a training set $\gY_{seen}$ corresponds to the subset of all possible pairs of state and object, $\gY_{seen} \subset \gY_{all}$, where $\gY_{all} = \{(s, o) | s \in \gS \,\,\text{and}\,\, o \in \gO \}$. The model is trained with seen candidate labels $\gY_{seen}$, and the goal of CZSL is to classify an image from the test label space $\gY_{test}$ that contains both seen and unseen classes, $\gY_{test} = \gY_{seen} \cup \gY_{unseen}$, where $\gY_{unseen}$ denotes the unseen classes during the training, $\gY_{seen} \cap \gY_{unseen} = \emptyset$ and $\gY_{unseen} \subset \gY_{all}$.

The open-world setting assumes no prior information about the set of unseen compositions at test time, i.e., $\gY_{test} \coloneqq \gY_{all}$. The substantial increase in the label space in the open-world setting leads to a significant performance gap compared to the closed-world setting where the label space is known. 
To mitigate this performance gap, previous works~\citep{owczsl, kgsp} have developed a function $g(\cdot)$ that assigns a feasibility score to each class such that classes with scores below a threshold $\tau$ are deemed infeasible and consequently removed from the test label space, $\gY_{test} \coloneqq \{y| y \in \gY_{all} \,\,\text{and}\,\, g(y) \ge \tau\}$.

An accurate feasibility function plays a critical role in open-world CZSL. If the function assigns low scores to feasible classes, the correct class is absent from the label space. Conversely, if the function assigns high feasibility to numerous infeasible classes, incorrect predictions become more likely.

\begin{table*}[t]
\centering
\setlength{\tabcolsep}{3pt}
\resizebox{0.8\textwidth}{!}{
\renewcommand{\arraystretch}{1.3}
\begin{tabular}{llcccc|cccc|cccc}
\toprule
\multicolumn{2}{l}{\bf ViT-L/14} & \multicolumn{4}{c|}{MIT-States} & \multicolumn{4}{c|}{UT-Zappos} & \multicolumn{4}{c}{C-GQA} \\ \hline
VLM & Method & S & U & H & AUC & S & U & H & AUC & S & U & H & AUC \\
\hline
\multirow{3}{*}{CLIP}
& GloVe
& 30.21 & 14.6 & 13.0 & 3.10
& 10.8 & 19.3 & 10.6 & 1.60 
& 7.59 & \bf 3.92 & 2.46 & \bf 0.20 \\
& ConceptNet
& \ditto & 12.5 & 12.7 & 2.75 
& \ditto & 21.3 & 10.3 & 1.69 
& \ditto & 2.01 & 2.59 & 0.13 \\
& \ours\ (ours)  
& \ditto & \bf 16.1 & \bf 13.7 & \bf 3.38
& \ditto & \bf 23.6 & \bf 11.5 & \bf 1.94
& \ditto & 2.62 & \bf 2.82 & 0.16  \\
\hline
\multirow{3}{*}{CoOp} 
& GloVe
& 38.2\scriptsize$\pm0.9$ & 16.7\scriptsize$\pm0.3$ & 16.2\scriptsize$\pm0.4$ & 4.78\scriptsize$\pm0.2$ 
& 61.2\scriptsize$\pm1.8$ & 36.7\scriptsize$\pm2.4$ & 34.2\scriptsize$\pm3.7$ & 18.1\scriptsize$\pm2.8$ 
& 26.9\scriptsize$\pm1.6$ & 5.09\scriptsize$\pm0.5$ & 6.16\scriptsize$\pm0.5$ & 1.03\scriptsize$\pm0.1$ \\
& ConceptNet
& \ditto & 14.5\scriptsize$\pm0.3$ & 15.6\scriptsize$\pm0.3$ & 4.36\scriptsize$\pm0.1$
& \ditto & 42.1\scriptsize$\pm2.3$ & 36.6\scriptsize$\pm3.2$ & 20.6\scriptsize$\pm2.5$
& \ditto & 4.14\scriptsize$\pm0.5$ & 5.88\scriptsize$\pm0.7$ & 0.92\scriptsize$\pm0.2$ \\
& \ours\ (ours)  
& \ditto & \bf 18.7\scriptsize$\pm0.3$ & \bf 17.4\scriptsize$\pm0.5$ & \bf 5.40\scriptsize$\pm0.1$ 
& \ditto & \bf 49.6\scriptsize$\pm1.7$ & \bf 40.6\scriptsize$\pm3.1$ & \bf 24.4\scriptsize$\pm2.7$
& \ditto & \bf 5.16\scriptsize$\pm0.3$ & \bf 6.91\scriptsize$\pm0.3$ & \bf 1.13\scriptsize$\pm0.1$ \\
\hline
\multirow{3}{*}{CSP} 
& GloVe
& 45.1\scriptsize$\pm0.9$ & 14.9\scriptsize$\pm0.3$ & 16.4\scriptsize$\pm0.4$ & 5.12\scriptsize$\pm0.2$
& 62.8\scriptsize$\pm0.9$ & 45.8\scriptsize$\pm1.8$ & 38.9\scriptsize$\pm0.8$ & 22.6\scriptsize$\pm1.0$
& 30.2\scriptsize$\pm0.5$ & \bf 4.58\scriptsize$\pm0.5$ & 6.12\scriptsize$\pm0.5$ & 1.09\scriptsize$\pm0.1$ \\
& ConceptNet 
& \ditto & 13.4\scriptsize$\pm0.8$ & 15.5\scriptsize$\pm0.5$ & 4.74\scriptsize$\pm0.3$ 
& \ditto & 54.0\scriptsize$\pm1.7$ & 43.3\scriptsize$\pm0.9$ & 26.9\scriptsize$\pm1.1$
& \ditto & 1.31\scriptsize$\pm0.1$ & 2.25\scriptsize$\pm0.3$ & 0.34\scriptsize$\pm0.0$ \\
& \ours\ (ours)  
& \ditto & \bf 16.6\scriptsize$\pm0.3$ & \bf 17.4\scriptsize$\pm0.6$ & \bf 5.76\scriptsize$\pm0.2$
& \ditto & \bf 56.7\scriptsize$\pm1.3$ & \bf 43.9\scriptsize$\pm0.9$ & \bf 30.0\scriptsize$\pm1.1$
& \ditto & 4.55\scriptsize$\pm0.5$ & \bf 6.55\scriptsize$\pm0.5$ & \bf 1.13\scriptsize$\pm0.1$  \\
\bottomrule
\end{tabular}
}
\vspace{-2mm}
\caption{
CSZL results comparing Glove, ConceptNet and our \ours\ (Vicuna, logit) on three benchmarks. We report seen (S) and unseen class accuracy (U), harmonic mean (H) and AUC using the CLIP, CoOp and CSP as base models. Ditto (\ditto) denotes ``same as above''.
}
\vspace{-5pt}
\label{table:main}
\end{table*}

\subsection{\oursFull\ (\ours)}
\label{subsec:method}

LLMs generate words autoregressively, i.e., they model $p_{\text{LLM}}(t_k|t_1,\dots,t_{k-1})$ where $t$ is a token from the vocabulary of the language model. The output probability indicates how certain the LLM is that a given token should appear next, which is used by \ours\ as a measure of feasibility.

\myparagraph{Canonical prompt.}
To obtain feasibility scores using LLMs, we construct a prompt $c$ that consists of a system message, \textit{sysmsg}, and a human message, \textit{hmsg}. The \textit{sysmsg} provides the LLM with general guidance while the \textit{hmsg} asks the LLM to assess the feasibility of a given class.
\begin{align*}
    c = \{\,
    sysmsg &: ``\,\texttt{\small You are a helpful, respectful} \\[-4pt] 
    \,&\quad\;\, \texttt{\small and honest assistant. Answer} \\[-4pt]
    \,&\quad\;\, \texttt{\small with a single word, yes or no.}\,", \\[-4pt]
    hmsg &: ``\,\texttt{\small Does a/an \{s\} \{o\} exist in the} \\[-4pt] 
    \,&\quad\;\, \texttt{\small real world?}\," \,\}    
\end{align*}
where we refer to the first sentence of the \textit{sysmsg} as the \textit{persona} component~\citep{salewski2023persona}, the second as the \textit{instruction} component, and the sentence of the \textit{hmsg} as the \textit{query} component.
The placeholders \{$s$\} and \{$o$\} represent the state and object of the class, respectively.

The probability or the logit of the word "Yes" in the next-token output distribution indicates the LLM's confidence in the feasibility of the given pair $(s,o)$. We interpret this output as a feasibility score. 
More formally, our feasibility score function is
\begin{equation}
    g(s,o) = \log p_{\text{LLM}}\left(t=\text{\texttt{"Yes"}}| f(s,o;c)\right)
\end{equation}
where $\log p_{\text{LLM}}$ indicates the unnormalized output logits and $f(s,o;c)$ denotes a function that composes the prompt $c$ with the target state-object pair $(s, o)$.
To obtain a real-valued score, this requires local access to the LLM. When an LLM is accessed through an API, such as ChatGPT, which does not expose the output probabilities or logits, we can only retrieve a binary score of "Yes" or "No".

\myparagraph{In-context learning.} 
Directly querying the LLM about the feasibility might result in incorrect responses, e.g. "dark fire" as mentioned in \S\ref{sec:intro}.
Motivated by this, we leverage the in-context learning capabilities of LLMs which allows them to adapt to new tasks given few examples. 
We introduce a \textit{guidance} component that includes a few examples of true feasible pairs, allowing the LLMs to learn from these instances and better understand what constitutes feasibility within the dataset. For example, a prompt with guidance would be:
\begin{align*}
    ``\, &\texttt{The following list consists of word} \\[-4pt] 
    &\texttt{combinations that make sense.} \\[-4pt]
    &\small - \; \{s_1\} \; \{o_1\} \\[-4pt]
    &\small - \; \cdots \\[-4pt]
    &\small - \; \{s_n\} \; \{o_n\} \\[-4pt]
    &\texttt{Does "\{$s$\}\;\{$o$\}" fit into the list above?}\,"
\end{align*}
where $\{(s_i, o_i)\}_{i=1}^n$ are few-shot examples of true feasible pairs.
Motivated by \cite{owczsl}, we choose guidance pairs from the seen classes that either include the state $s$ or the object $o$, i.e., $\gY_{pos} = \{ (s_i, o_i) | (s_i, o_i) \in \gY_{seen},\; s_i = s \;\text{or}\; o_i = o \}$. This strategy enables the LLMs to gain a deeper understanding of the dataset-specific task within the in-context learning framework, improving predictions of the feasibility of query pairs. The overall pipeline is drawn in Figure~\ref{fig:main}. 
Our feasibility score function, denoted as \oursFull\ (\ours), is formulated as
\begin{equation}
    g(s,o) = \log p_{\text{LLM}}\left(t=\text{\texttt{"Yes"}}| f(s,o,\gY_{pos};c)\right)
\end{equation}
where $f(s,o,\gY_{pos};c)$ denotes a function that composes the prompt $c$ with the target state-object pair $(s, o)$ and the related seen pairs $\gY_{pos}$.

\myparagraph{Versatility.} Once we obtain the feasibility scores for all combinations of pairs, a threshold $\tau$ determines the subset of all pairs that are deemed feasible.
The infeasible pairs are discarded, and only the feasibile pairs are used as candidate labels for the VLM's prediction. Our feasibility scores can be integrated with any existing VLM to improve performance in the open-world setting.

\section{Experiments}

We present our experimental findings as well as ablations on LLM-guided feasibility prediction in OW-CZSL. Additional results and details about the implementation, benchmarks, metrics, and baselines are described in the Appendix.

\begin{table*}[t]
\centering
\setlength{\tabcolsep}{3pt}
\resizebox{.7\textwidth}{!}{
\renewcommand{\arraystretch}{1.0}
\begin{tabular}{lcccc|cccc|cccc}
\toprule
& \multicolumn{4}{c|}{MIT-States} & \multicolumn{4}{c|}{UT-Zappos} & \multicolumn{4}{c}{C-GQA} \\ \hline
& 
\begin{tabular}[c]{@{}c@{}}Feas.\\acc.\end{tabular} & 
\begin{tabular}[c]{@{}c@{}}Infeas.\\acc.\end{tabular} & 
\begin{tabular}[c]{@{}c@{}}Arith.\\mean\end{tabular} & 
\begin{tabular}[c]{@{}c@{}}H.\\mean\end{tabular} & 
\begin{tabular}[c]{@{}c@{}}Feas.\\acc.\end{tabular} & 
\begin{tabular}[c]{@{}c@{}}Infeas.\\acc.\end{tabular} & 
\begin{tabular}[c]{@{}c@{}}Arith.\\mean\end{tabular} & 
\begin{tabular}[c]{@{}c@{}}H.\\mean\end{tabular} & 
\begin{tabular}[c]{@{}c@{}}Feas.\\acc.\end{tabular} & 
\begin{tabular}[c]{@{}c@{}}Infeas.\\acc.\end{tabular} & 
\begin{tabular}[c]{@{}c@{}}Arith.\\mean\end{tabular} & 
\begin{tabular}[c]{@{}c@{}}H.\\mean\end{tabular} \\
\hline
GloVe
 & 51.7
 & \bf 93.2
 & 72.5
 & 66.5
 & 78.8
 & 38.2
 & 58.5
 & 51.4
 & 40.0
 & \bf 98.5
 & 69.2
 & 56.9 \\
ConceptNet
 & 52.0
 & 92.5
 & 72.3
 & 66.6
 & \bf 100.0
 & 13.2
 & 56.6
 & 23.3
 & 26.3
 & 91.7
 & 59.0
 & 40.9 \\
\ours\ (ours)
 & \bf 64.7
 & 86.1
 & \bf 75.4
 & \bf 73.9
 & 93.9
 & \bf 46.1
 & \bf 70.0
 & \bf 61.8
 & \bf 70.9
 & 89.2
 & \bf 80.1
 & \bf 79.0 \\
\bottomrule
\end{tabular}
}
\vspace{-2mm}
\caption{Accuracy of correctly identifying feasible and infeasible open-world pairs from $\gY_{all}$ using the same threshold $\tau$ as in Table~\ref{table:main}. 
}
\vspace{-10pt}
\label{table:feasibility}
\end{table*}

\subsection{Experimental setup}
\label{subsec:experimental_setup}

\myparagraph{Benchmarks.} We use three standard datasets for OW-CZSL, i.e., MIT-States~\citep{mitstates}, UT-Zappos~\citep{utzappos1,utzappos2}, and C-GQA~\citep{cgqa}.

\myparagraph{Evaluation metric.} 
We follow the protocol of \cite{purushwalkam2019task} for OW-CZSL which has 4 metrics: seen class accuracy (denoted as S), unseen class accuracy (U), harmonic mean of accuracy on seen class and unseen class (H), and area under the curve of seen class and unseen class accuracy (AUC).

\myparagraph{Feasibility baselines.} We compare with GloVe embeddings~\citep{glove} as used in CompCos~\citep{owczsl} and CSP~\citep{csp}, and the ConceptNet~\citep{conceptnet} as used in KGSP~\citep{kgsp}. For the LLM, we use the Vicuna-13B model~\citep{vicuna} for the experiments unless otherwise indicated.

\subsection{LLM-generated feasibility in OW-CZSL}
\label{subsec:experiment_main_result}

We evaluate our \ours\ method using three VLMs: CLIP~\citep{clip}, CoOp~\citep{coop}, and CSP~\citep{csp}, which are CLIP-based models. We choose VLMs since they outperform the CNN-based task-specific CZSL methods, as reported by \cite{csp}.

The experimental results are presented in Table \ref{table:main} where we improve OW-CZSL performance on all datasets.
We observe that on the MIT-States dataset, our \ours\ method achieves the highest harmonic mean of 17.4\% and AUC of 5.76\%, surpassing the GloVe feasibility function which shows 16.4\% and 5.12\%, and the ConceptNet feasibility function with 15.5\% and 4.74\% using the CSP model.
Overall, \ours\ achieves the best results on all metrics across all datasets. These results indicate that \ours\ can better differentiate between feasible state-object pairs and infeasible ones, as it facilitates all base OW-CZSL models to obtain a higher score, closing the gap to the closed-world setting.

\subsection{
\add{Ablation Study}
}
\label{subsec:feasibility}

\myparagraph{Feasibility Prediction in Isolation from the OW-CZSL Task.}
We evaluate overlap of the feasibility scores with the human annotations quantitatively in Table~\ref{table:feasibility}. For every state-object pair in $\gY_{all}$, we compute the feasibility prediction of GloVe, ConceptNet, and \ours\ and compare it to the human-annotated ground truth of feasible classes $\gY_{unseen}$. We report feasibility accuracy as the ratio of pairs in $\gY_{unseen}$ correctly identified as feasible and, analogously, infeasible accuracy for the ratio of all other classes predicted as infeasible.
We observe that our \ours\ performs the best on either feasible accuracy or infeasible accuracy, which suggests that the best trade-off between feasible and infeasible accuracy varies by dataset. Considering both metrics together through arithmetic and harmonic means, our method performs the best across all datasets, often by a significant margin.

\begin{figure}[t]
    \centering
    \includegraphics[width=\linewidth]{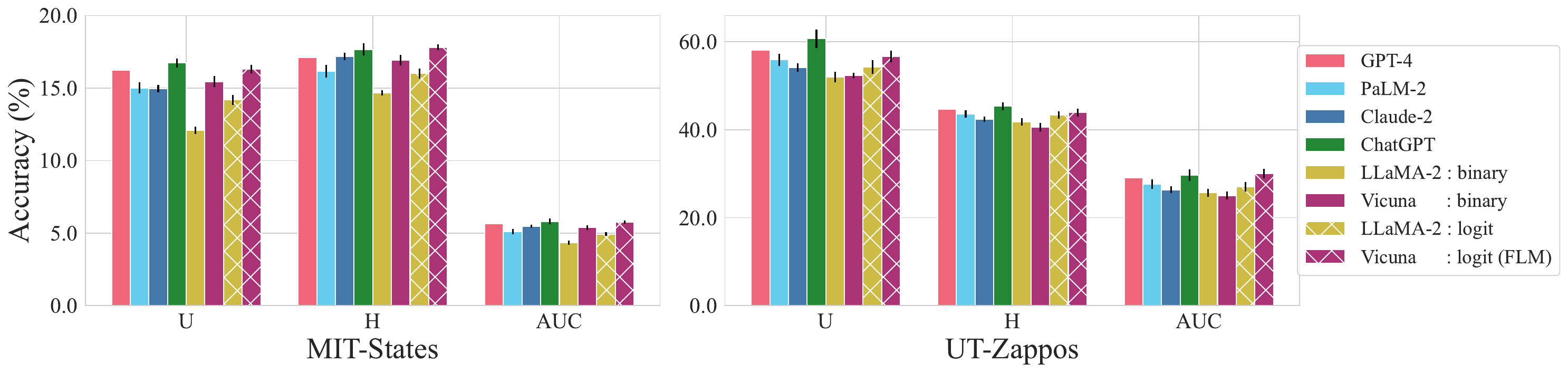}
    \vspace{-20pt}
    \caption{
    Comparison of \ours\ using Vicuna, LLaMA-2, and proprietary models (GPT-4, PaLM-2, Claude-2, ChatGPT) as LLMs on MIT-States and UT-Zappos. Proprietary models can only provide a binary ``Yes'' or ``No'' response, whereas for Vicuna and LLaMA-2 we evaluate both the binary and logit outputs as feasibility scores. 
    }
    \vspace{-10pt}
    \label{fig:chatgpt_comparison}
\end{figure}

\myparagraph{Comparing Large Language Models.} We use ChatGPT~\citep{instructgpt,gpt3}, GPT-4~\cite{gpt4}, PaLM-2~\cite{palm2}, Claude-2~\citep{claude2}, LLaMa-2-Chat-13B~\citep{llama2} and Vicuna-13B~\citep{vicuna}. One advantage of Vicuna and LLaMa-2 over proprietary models is the accessibility of internal values of the output words. We utilize the logit value of the word ``Yes'' as our feasibility score.
Moreover, we compare with ChatGPT, GPT-4, PaLM-2 and Claude-2 as proprietary LLMs where we query the API to obtain binary feasibility scores, i.e., a score of 1 when the model answers with ``Yes'' and 0 when it answers with ``No''. Due to API constraints, we conduct these experiments only on the MIT-States and UT-Zappos datasets as shown in Figure \ref{fig:chatgpt_comparison}. We observe that 1) both Vicuna and LLaMa-2 show lower performance with a binary answer compared to using logits, 2) Vicuna clearly outperforms LLaMa-2, 3) among the proprietary LLMs, ChatGPT performs best, and 4) ChatGPT with a binary answer consistently outperforms Vicuna with a binary answer and oftentimes even Vicuna using logits. From these findings, we speculate that ChatGPT with logit access would likely surpass Vicuna with logit considerably, implying that more advanced LLMs with logit access would yield improved feasibility scores and, thus, could further push state-of-the-art in OW-CZSL.

\section{Conclusion}

We proposed a novel approach that leverages LLMs to predict the feasibility of the state-object pair for the open-world compositional zero-shot learning (OW-CZSL). We designed prompts to query the feasibility of class pairs to LLMs by considering the output of the word ``Yes'' as feasibility score. We used the in-context learning capabilities of LLMs by providing guidance prompts that included a few examples of true feasible pairs. Our experimental results validated the effectiveness of our \ours\ which achieves better performance than previous approaches on three standard benchmarks.

\newpage

\section*{Acknowledgements}

Jae Myung Kim thanks the International Max Planck Research School for Intelligent Systems (IMPRS-IS)
and the European Laboratory for Learning and Intelligent Systems (ELLIS) PhD programs for support. This work was supported by the ERC (853489 - DEXIM) and the Alfried Krupp von Bohlen und Halbach Foundation, which we thank for their generous support.
Cordelia Schmid would like to acknowledge the support by the K\"orber European Science Prize. 
The authors gratefully acknowledge the Gauss Centre for Supercomputing e.V. (www.gauss-centre.eu) for funding this project by providing computing time on the GCS Supercomputer JUWELS at Jülich Supercomputing Centre (JSC).

{
    \small
    \bibliographystyle{ieeenat_fullname}
    \bibliography{main}
}

\clearpage
\appendix
\section*{Appendix}
\renewcommand\thesection{\Alph{section}}

\section{Broader Impact and Limitations}
Determining the feasibility of state-object pairs in all combinations is crucial when deploying the model. By accurately assessing feasibility, we prevent the model from predicting unrealistic classes, thus improving the model performance and reducing negative impacts on end-users. However, our \ours\ is limited in that we use prior knowledge, i.e. seen classes from the training dataset, which could be biased. If the seen classes represent only part of the semantics, e.g. only the texture-related attributes like ``furry'' or ``zigzag'' for animal in the seen classes while age-related states exist in the test set, our method may struggle to predict the true feasible pairs accurately. The bias can also lead to fairness issues if the seen classes only represent certain groups. Therefore, it is important to curate an unbiased seen class set to support the model's understanding of feasibility for the query pairs.

\section{Prompt search in \ours} 
%
There are four prompt components in \ours: persona, instruction, guidance, and query. In our experiments, We keep the \textit{persona} component fixed as ``\texttt{\small You are a helpful, respectful and honest assistant.}''. 
We conduct a grid search using four instruction, four guidance, and four query sentences, which are:
%
{\small
\begin{align*}
    &instruction\_list = \{ \\
    &\qquad ``\,\texttt{Answer with a single word, yes or no.}\,", \\
    &\qquad ``\,\texttt{Answer with a single word, yes or no, followed by an explanation.}\,", \\
    &\qquad ``\,\texttt{Answer with yes or no.}\,", \\
    &\qquad ``\,\texttt{Answer with yes or no, followed by an explanation.}\,", \\
    &\}, \\
    &guidance\_list = \{ \\
    &\qquad ``\,\texttt{The following list consists of words that fit together.}\,", \\
    &\qquad ``\,\texttt{The following list consists of word combinations that make sense.}\,", \\
    &\qquad ``\,\texttt{The given list consists of word combinations that make sense.}\,", \\
    &\qquad ``\,\texttt{The given list comprises word combinations that make sense.}\,", \\
    &\}, \\
    &query\_list = \{ \\
    &\qquad ``\,\texttt{Considering the list above, does "\{\textit{s}\}\;\{\textit{o}\}" fit into the list?}\,", \\
    &\qquad ``\,\texttt{Does "\{\textit{s}\}\;\{\textit{o}\}" fit into the list above?}\,", \\
    &\qquad ``\,\texttt{Does "\{\textit{s}\}\;\{\textit{o}\}" align with the contents of the list provided above?}\,", \\
    &\qquad ``\,\texttt{Considering the list above, does "\{\textit{s}\}\;\{\textit{o}\}" align with the contents?}\,", \\
    &\} \, ,
\end{align*}
}
and select the combination that yields the highest unseen validation accuracy following the validation protocol and split of~\cite{csp}. On MIT-States, we use ``\texttt{\small Answer with a single word, yes or no, followed by an explanation.}'' as instruction, ``\texttt{\small The following list consists of words that fit together.}'' as guidance, and ``\texttt{\small Does "\{$s$\}\;\{$o$\}" fit into the list above?}'' as query. On the UT-Zappos, we use ``\texttt{\small Answer with a single word, yes or no.}'' as instruction, ``\texttt{\small The given list consists of word combinations that make sense.}'' as guidance, and ``\texttt{\small Considering the list above, does "\{\textit{s}\}\;\{\textit{o}\}" fit into the list?}'' as query. Finally on the C-GQA, we use ``\texttt{\small Answer with a single word, yes or no, followed by an explanation.}'' as instruction, ``\texttt{\small The given list consists of word combinations that make sense.}'' for CLIP and ``\texttt{\small The given list comprises word combinations that make sense.}'' for CoOp and CSP as guidance, and ``\texttt{\small Does "\{\textit{s}\}\;\{\textit{o}\}" align with the contents of the list provided above?}'' as query.

\clearpage

\begin{figure}[t]
    \centering
    \vspace{-5pt}
    \includegraphics[width=\linewidth]{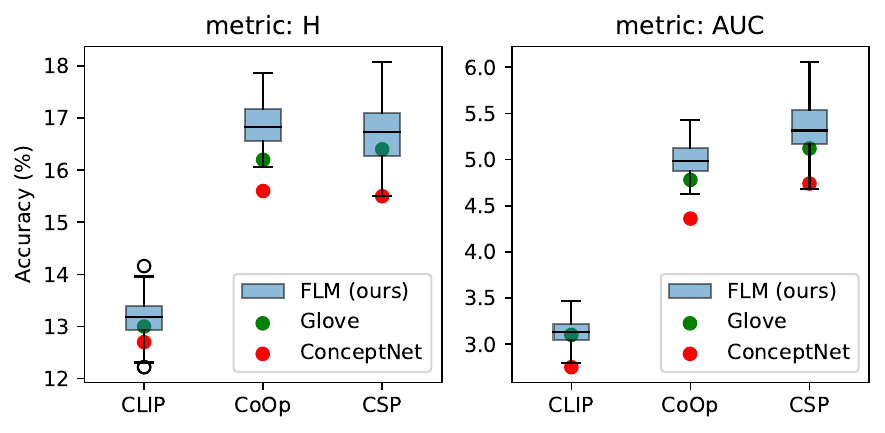} 
    \vspace{-15pt}
    \caption{Prompt variation results on MIT-States dataset.}
    \label{fig:prompt_variation}
\end{figure} 
To examine a broader range of prompt variations, we report the results on MIT-States as a box plot in Figure~\ref{fig:prompt_variation}. We first observe the results vary with different prompts. For instance, the Harmonic mean accuracy in CSP ranges from 15.5\% to 18.1\%. However, despite this variability, most of the prompts outperform the baselines. For instance, the Glove result always lies close to or below the lower quartile (25th percentile) of the box plot, and ConceptNet even further below that.

\section{Implementation Details}

\subsection{Datasets}
Each dataset comprises a set of states and objects, where an object-state combination forms a class. MIT-States consists of 115 states and 245 objects, resulting in a total of 28,175 possible pairs. Among these possible pairs, 1,262 and 700 pairs are seen classes and unseen classes, respectively. UT-Zappos includes 16 states and 12 objects, leading to 192 possible pairs, with 83 seen classes and 33 unseen classes. Finally, C-GQA has 413 states and 674 objects, resulting in 278,362 possible pairs, with 5,592 seen classes and 1,963 unseen classes.

\subsection{Evaluation Metrics}
We follow the protocol of \cite{purushwalkam2019task} for OW-CZSL. Since the VLM is trained only on seen classes, it is prone to being biased towards classifying an image as one of the seen classes at test time.
Concretely, a calibration bias is subtracted from the model outputs of the seen classes, and then the class is predicted. The calibration bias is varied to get the best combination of seen class accuracy (denoted as S), unseen class accuracy (U), harmonic mean of accuracy on seen class and unseen class (H), and area under the curve of seen class and unseen class accuracy (AUC). By tackling the feasibility prediction of unseen classes, we focus on improving the more challenging metrics (U, H, AUC) while the seen class accuracy (S) remains unaffected.

\subsection{Hyperparameter settings in CZSL models}
For OW-CZSL, hyperparameters are traditionally explored, and the best model is chosen based on the highest unseen validation accuracy. We perform a grid search on sentence variations as well as choose the threshold $\tau$ that determines whether the query class is feasible by the best unseen validation accuracy.
We train CoOp~\citep{coop} and CSP~\citep{csp} with the CSP official code~\footnote{\url{https://github.com/BatsResearch/csp}} to get the fine-tuned models. Following the original CSP setting, we fine-tune a pretrained CLIP~\citep{clip} model with a ViT-L/14 backbone for 20 epochs and choose the checkpoint with the highest validation accuracy on unseen classes. During training, we employ a batch size of 64, a learning rate of 5e-4, and a weight decay of 1e-5. Additionally, we set the attribute dropout rate for CSP to 0.3. We did not optimize these hyperparameters as originally done by \cite{csp}, although optimizing these hyperparameters as done by \cite{csp} could further yield improvements. We run CoOp and CSP with 5 different seeds and report the mean and the standard deviation. The model training is performed on a single A100 GPU. Similarly, when querying Vicuna-13B for our \ours, we use a single A100 GPU.


\subsection{Baselines}
For GloVe, the cosine similarity between the concepts of the same primitives are calculated and merged to represent the feasibility score. ConceptNet~\citep{conceptnet} is a knowledge graph connecting words to obtain the feasibility scores which are calculated by the cosine similarities of ConceptNet embeddings.


\section{Question-Answer format: 0-9 score}

\begin{wrapfigure}{r}{4.5cm}
    \centering
    \vspace{-10pt}
    \includegraphics[width=\linewidth]{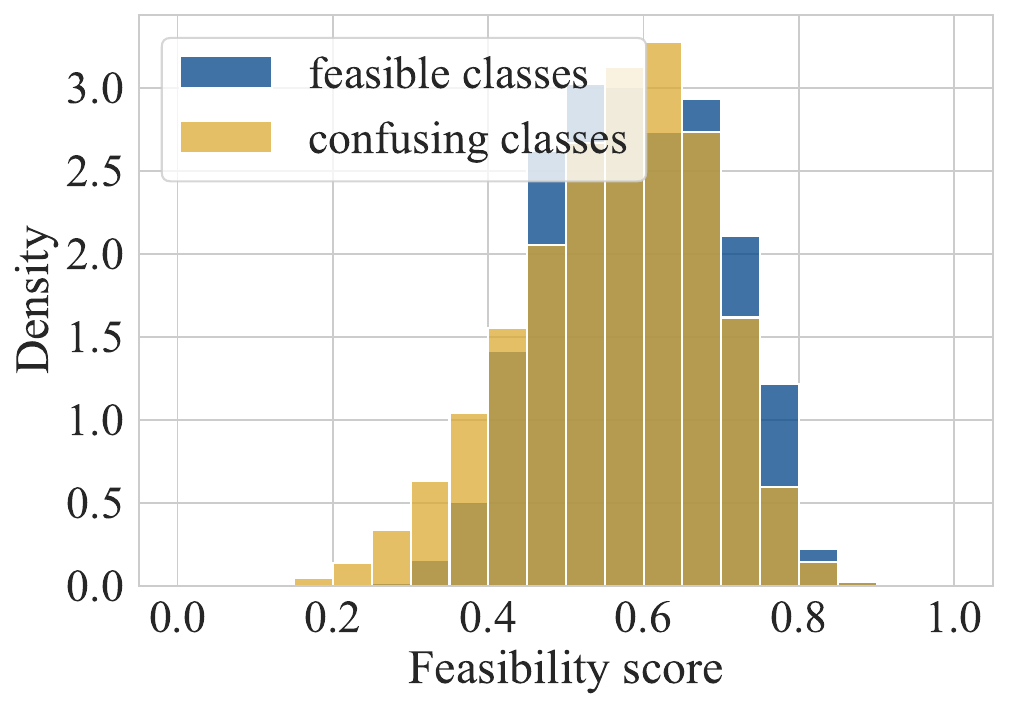} 
    \vspace{-15pt}
    \caption{Feasibility scores for question-answer ``yes'' format in the in-context prompt.}
    \vspace{-10pt}
    \label{fig:qa_ablation}
\end{wrapfigure} 
As mentioned in the main text, we observe that the prompt ablation ``Format QA: yes'' drops performance across all datasets, e.g. 17.4\%$\rightarrow$12.0 and 6.55\%$\rightarrow$3.69\% in harmonic mean on MIT-States and C-GQA. The lower performance originates from the tendency of the LLM to answer "Yes" since the provided examples in the guidance are always positive. This phenomenon is also evident in Figure~\ref{fig:qa_ablation}, where both the distributions of feasible and confusing classes overlap and lean closer towards 1. These observations indicate that employing a list format for guidance rather than a question-answer format is crucial in obtaining accurate feasibility scores. 

Similar trends are observed for ``Format QA: score'' where we use a question-answer format with guiding LLMs to respond with an integer score instead of a binary answer. To obtain an integer score as an answer from the LLMs, we construct the guidance component of the human message as:
{\small 
\begin{align*}
    ``\, &\texttt{The following list consists of words and} \\[-4pt]
    &\texttt{their likelihood of existence in the real} \\[-4pt]
    &\texttt{world, scored on a scale of 0 to 9.} \\[-4pt]
    &\small - \; \{s_1\} \; \{o_1\}, \texttt{ score: 9} \\[-4pt]
    &\small - \; \{s_2\} \; \{o_2\}, \texttt{ score: 9} \\[-4pt]
    &\small - \; \cdots \\[-4pt]
    &\small - \; \{s_n\} \; \{o_n\}, \texttt{ score: 9} \\[-4pt]
    &\texttt{What is the score for "\{$s$\}\;\{$o$\}"?}\,"
\end{align*}
}
%
If we had access to a more nuanced classification of prior feasibility scores for the seen classes, we could provide the LLM with a more informative guidance. As this is not available, we had to choose fixed score value.
The results are shown in the ``Format QA: score'' row of Table \ref{table:ablation_prompt} in the main text. We observe the performance drops across the datasets, e.g. 17.4\%$\rightarrow$12.3\% and 6.55\%$\rightarrow$3.69\% in harmonic mean on MIT-States and C-GQA.
These observations, together with ``QA: yes'' format, indicate that employing a list format for guidance rather than a question-answer format is crucial in obtaining accurate feasibility scores. 

\section{Additional ablation study}
\subsection{Feasibility Prediction in Isolation from the OW-CZSL Task.}

\begin{figure*}[t]
    \centering
    \vspace{5pt}
    \includegraphics[width=\linewidth]{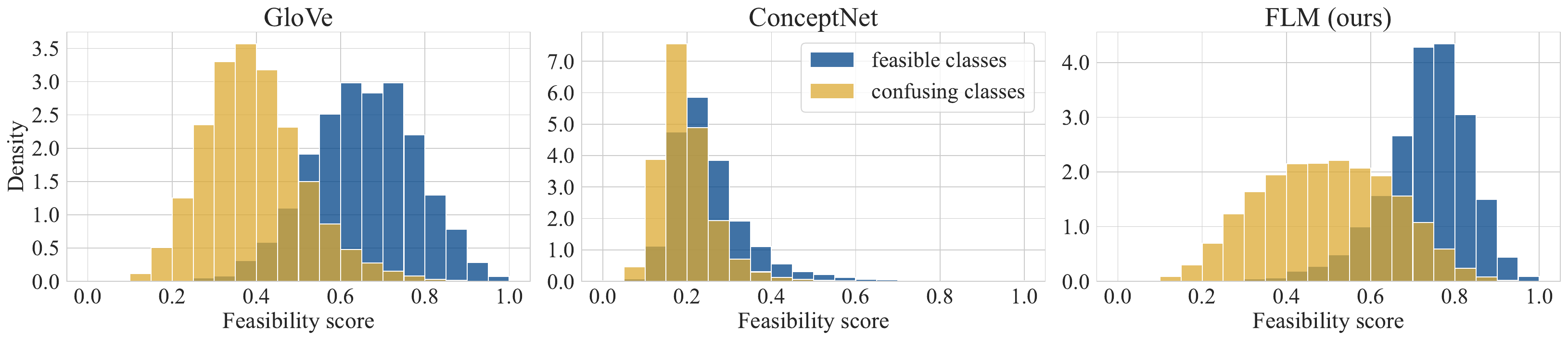} 
    \caption{Distributions of feasibility scores of all state-object pairs. For best separation, feasible classes should be close to 1 and all remaining confusing classes close to 0. 
    }
    \label{fig:dist}
\end{figure*}

We evaluate the feasiblity prediction in isolation from the OW-CZSL task and analyze the distributions of feasibility scores on the C-GQA dataset. In
Figure~\ref{fig:dist}, the unseen classes are referred to as ``feasible classes'' (blue) and the classes that are absent in both the seen and unseen sets are referred to as ``confusing classes'' (orange). Note that the scores obtained from each method have been normalized to fall within the range of 0 and 1. \ours\ exhibits a better separation between the two distributions than GloVe and ConceptNet implying that our approach more effectively distinguishes between feasible and infeasible classes, providing an accurate assessment of the feasibility. Quantitative analysis is reported in the Appendix.

\subsection{Ablation Study: Analysis of LLM-Prompts}
\label{subsec:analysis}

\begin{table*}[t]
\centering
\setlength{\tabcolsep}{3pt}
\resizebox{\textwidth}{!}{
\renewcommand{\arraystretch}{1.3}
\begin{tabular}{llccc|ccc|ccc}
\toprule
& & \multicolumn{3}{c|}{MIT-States} & \multicolumn{3}{c|}{UT-Zappos} & \multicolumn{3}{c}{C-GQA} \\ \hline
Prompt & & U & H & AUC & U & H & AUC & U & H & AUC \\
\hline
\multicolumn{2}{l}{Canonical}
& 13.5\scriptsize$\pm0.4$ & 15.4\scriptsize$\pm0.3$ & 4.70\scriptsize$\pm0.2$ 
& 47.1\scriptsize$\pm1.4$ & 39.2\scriptsize$\pm0.6$ & 23.1\scriptsize$\pm0.6$  
& 2.16\scriptsize$\pm0.5$ & 3.40\scriptsize$\pm0.6$ & 0.52\scriptsize$\pm0.1$  \\
\multicolumn{2}{l}{Instruction: hmsg begin}
& 15.5\scriptsize$\pm0.4$ & 16.6\scriptsize$\pm0.3$ & 5.32\scriptsize$\pm0.2$ 
& 58.2\scriptsize$\pm1.2$ & 44.1\scriptsize$\pm0.6$ & 28.3\scriptsize$\pm0.7$ 
& 3.06\scriptsize$\pm0.4$ & 4.42\scriptsize$\pm0.3$ & 0.69\scriptsize$\pm0.0$ \\
\multicolumn{2}{l}{Instruction: hmsg last}
& 14.8\scriptsize$\pm0.3$ & 16.4\scriptsize$\pm0.3$ & 5.11\scriptsize$\pm0.1$ 
& 56.6\scriptsize$\pm1.2$ & 43.8\scriptsize$\pm0.9$ & 27.9\scriptsize$\pm1.1$ 
& 2.84\scriptsize$\pm0.1$ & 4.23\scriptsize$\pm0.3$ & 0.66\scriptsize$\pm0.1$ \\
\hline
\multirow{2}{*}{Format}   
& QA: yes
& 10.3\scriptsize$\pm0.7$ & 12.0\scriptsize$\pm0.5$ & 3.39\scriptsize$\pm0.3$ 
& 53.1\scriptsize$\pm1.6$ & 42.4\scriptsize$\pm1.3$ & 26.7\scriptsize$\pm1.2$ 
& 2.23\scriptsize$\pm0.3$ & 3.40\scriptsize$\pm0.6$ & 0.49\scriptsize$\pm0.1$ \\      
& QA: score
& 10.8\scriptsize$\pm0.3$ & 12.3\scriptsize$\pm0.3$ & 3.53\scriptsize$\pm0.1$ 
& 54.2\scriptsize$\pm1.6$ & 43.4\scriptsize$\pm0.9$ & 26.9\scriptsize$\pm1.0$ 
& 2.36\scriptsize$\pm0.1$ & 3.69\scriptsize$\pm0.1$ & 0.54\scriptsize$\pm0.0$ \\      
\hline
\multirow{5}{*}{In-context}
& $\gY_{pos}$, 5
& 13.8\scriptsize$\pm0.3$ & 15.3\scriptsize$\pm0.3$ & 4.74\scriptsize$\pm0.1$ 
& 48.2\scriptsize$\pm8.0$ & 39.9\scriptsize$\pm4.0$ & 24.0\scriptsize$\pm4.2$
& 3.16\scriptsize$\pm0.3$ & 4.67\scriptsize$\pm0.6$ & 0.76\scriptsize$\pm0.1$ \\   
& $\gY_{pos}$, 20
& 14.7\scriptsize$\pm0.6$ & 16.3\scriptsize$\pm0.3$ & 5.12\scriptsize$\pm0.1$ 
& 56.7\scriptsize$\pm1.3$ & 43.9\scriptsize$\pm0.9$ & 30.0\scriptsize$\pm1.1$ 
& 2.66\scriptsize$\pm0.5$ & 4.00\scriptsize$\pm0.7$ & 0.63\scriptsize$\pm0.1$ \\   
& $\gY_{pos}$, 50
& 16.6\scriptsize$\pm0.3$ & 17.4\scriptsize$\pm0.6$ & 5.76\scriptsize$\pm0.1$ 
& \ditto & \ditto & \ditto 
& 3.25\scriptsize$\pm0.5$ & 4.62\scriptsize$\pm0.7$ & 0.76\scriptsize$\pm0.1$ \\ 
& $\gY_{pos}$, 200
& \ditto & \ditto & \ditto 
& \ditto & \ditto & \ditto
& 4.55\scriptsize$\pm0.5$ &  6.55\scriptsize$\pm0.5$ &  1.13\scriptsize$\pm0.1$ \\  
& random, 200
& 13.2\scriptsize$\pm0.6$ & 14.9\scriptsize$\pm0.7$ & 4.50\scriptsize$\pm0.3$ 
& 50.4\scriptsize$\pm0.9$ & 40.7\scriptsize$\pm0.7$ & 24.8\scriptsize$\pm0.7$ 
& 2.95\scriptsize$\pm0.3$ & 4.51\scriptsize$\pm0.6$ & 0.72\scriptsize$\pm0.1$ \\ 
\Xhline{2\arrayrulewidth}
\multicolumn{2}{l}{\ours\ (ours)}
& \bf 16.6\scriptsize$\pm0.3$ & \bf 17.4\scriptsize$\pm0.6$ & \bf 5.76\scriptsize$\pm0.1$
& \bf 56.7\scriptsize$\pm1.3$ & \bf 43.9\scriptsize$\pm0.9$ & \bf 30.0\scriptsize$\pm1.1$ 
& \bf 4.55\scriptsize$\pm0.5$ & \bf 6.55\scriptsize$\pm0.5$ & \bf 1.13\scriptsize$\pm0.1$ \\
\bottomrule
\end{tabular}
}
\caption{ 
Ablation study. Experiments are done with CSP as VLM model. We ablate the canonical query without in-context exemplars, placing instruction component in the human message, a different format of the guidance component, varying the number of in-context pairs, and a random ordering of in-context examples. Ditto (\ditto) denotes the result is the same as the previous line. \delete{move this table to Appendix.}
}
\vspace{-10pt}
\label{table:ablation_prompt}
\end{table*}

\myparagraph{Comparison of instruction prompts.}
We investigate the impact of the prompt by comparing the performance of our in-context learning prompt with two ablations: 1) the canonical prompt described in \S\ref{subsec:method} which does not use in-context examples, and 2) placing instruction component, e.g. ``\texttt{\small Answer with a single word, yes or no.}'', in the human message instead of the system message. The results on the three datasets are presented in the first three rows of Table \ref{table:ablation_prompt}.

Across all datasets, we observe that using the canonical prompt significantly drops the performance, e.g. 17.4\%$\rightarrow$15.4\%, 43.9\%$\rightarrow$39.2\%, 6.55\%$\rightarrow$3.40\% in harmonic mean on MIT-States, UT-Zappos, and C-GQA, respectively. This highlights the importance of providing in-context guidance in our \ours\ for feasibility prediction. Moreover, placing an instruction in the human message, whether at the beginning or end, drops the performance on MIT-States and C-GQA while showing similar performance on UT-Zappos. This suggests that incorporating an instruction in the system message effectively guides the LLMs to the desired behavior.

\myparagraph{Format for in-context learning.} 
\cite{arora2022ask} have demonstrated the effectiveness of LLMs' in-context learning when using a question-and-answer format. To investigate this approach for \ours, we use a question-answer format for the guidance prompt instead of a list. Specifically, we employ the guidance prompt "\texttt{Does a/an \{$s_i$\} \{$o_i$\} exist in the real world? Yes.}", which is repeated for every related seen pair in $\gY_{pos}$, and the query prompt \texttt{"Does a/an \{$s$\} \{$o$\} exist in the real world?"} while keeping the rest of the process the same. The results are shown in the ``Format QA: yes'' row of Table \ref{table:ablation_prompt}. 

We observe the performance drops across datasets, e.g. 17.4\%$\rightarrow$12.0 and 6.55\%$\rightarrow$3.69\% in harmonic mean on MIT-States and C-GQA. The lower performance originates from this prompt format biasing the LLM to answer "Yes" because we only have access to feasible pairs.
We observe a similar trend for ``Format QA: score'' where we use the same question-answer format, but instruct the LLM to respond with an integer score indicating the level of feasibility (see Appendix for details).
Both of these results indicate that employing our proposed list format is crucial in obtaining accurate feasibility scores because we do not have access to infeasible examples.

\myparagraph{Number of pairs for in-context guidance.}
To analyze the influence of the number of in-context examples in the guidance prompt, we conducted experiments varying the number of positive pairs. Recall that our \ours\ method selects related pairs in the guidance as $\gY_{pos} = \{ (s_i, o_i) | (s_i, o_i) \in \gY_{seen},\; s_i = s \;\text{or}\; o_i = o \}$. The performance results are presented in the ``in-context'' rows of Table \ref{table:ablation_prompt}.

Across all datasets and evaluation metrics, performance consistently improves as the number of pairs in the guidance increases. For instance, on MIT-States, the harmonic mean increases from 15.3\% to 16.3\%, and subsequently to 17.8\%, as the number of pairs expands from 5, to 20, and to 50, respectively. Each dataset contains a different maximum number of related seen pairs. Thus, performance does not improve beyond 50 for MIT-States and beyond 20 for UT-Zappos.
Moreover, using up to 200 randomly selected state-object pairs results in worse performance than just 5 related pairs from $\gY_{pos}$ on MIT-States and C-GQA.
This suggests that it is important to provide relevant in-context pairs and that more few-shot examples allow the LLM to better comprehend the context-dependent task, leading to more accurate feasibility scores.

\section{Qualitative examples.}

\begin{figure*}[t]
    \centering
    \vspace{5pt}
    \includegraphics[width=\linewidth]{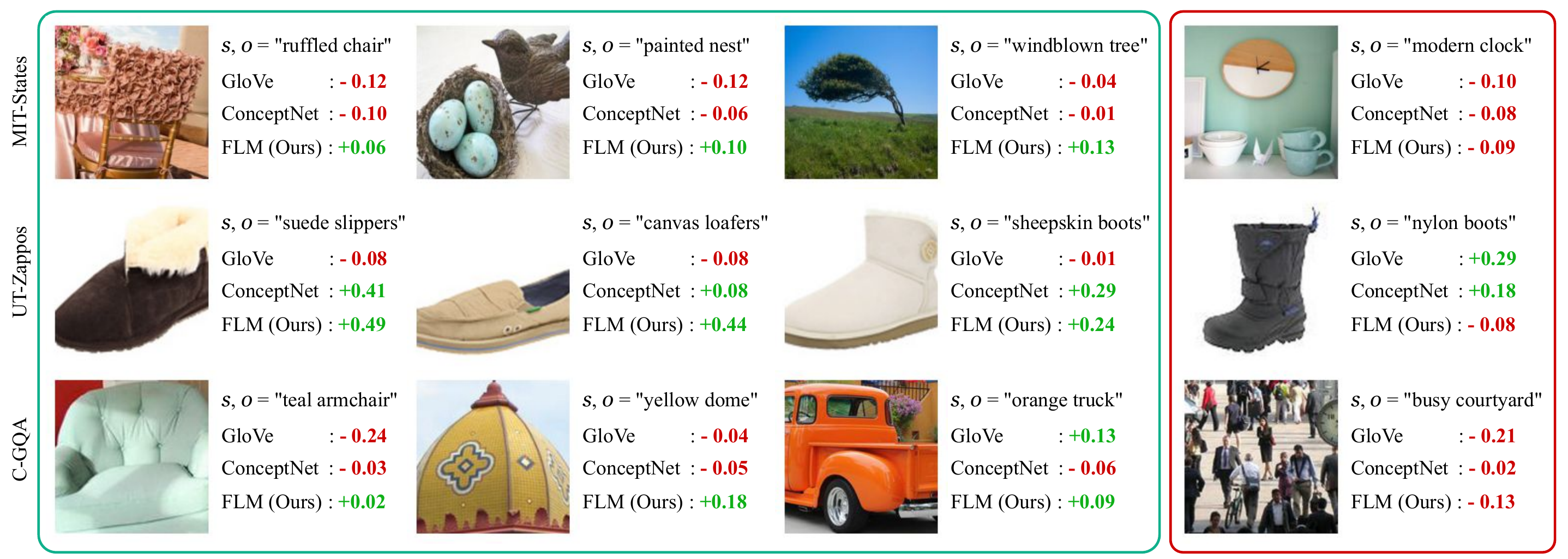} 
    \caption{
    Feasible examples from the unseen test set along with feasibility scores normalized such that the threshold $\tau$ is at 0. Positive scores (green) indicate a correct prediction as feasible, while negative scores (red) incorrectly infer infeasibility of a pair. Red box includes failure cases of \ours.
    }
    \label{fig:qualitative_with_images}
\end{figure*}

In Figure~\ref{fig:qualitative_with_images}, we show qualitative results of feasible images from the unseen classes alongside the absolute difference of the feasibility score from the threshold. 
\delete{
Positive values (green) indicate a correctly identified pair, while negative values (red) indicate an incorrect feasibility prediction. For each dataset, we show examples comparing \ours\ with GloVe and ConceptNet.
}
For instance, \ours\ correctly identifies that ``ruffled chair'' is feasible for the MIT-States dataset, and that ``teal armchair'' is feasible in the context of the C-GQA dataset, both of which are considered infeasible by GloVe and ConceptNet. By providing seen pairs that are relevant to the query pair in the guidance prompt, e.g. ``ruffled bed'' for the query ``ruffled chair'' and ``tan armchair'' for the query ``teal armchair'', our \ours\ correctly identifies the given query pairs as feasible.

\begin{table*}[h]
\centering
\setlength{\tabcolsep}{3pt}
\resizebox{\textwidth}{!}{
\renewcommand{\arraystretch}{1.3}
\begin{tabular}{l|ccccl|ccc}
\toprule
MIT-States & \multicolumn{3}{c}{Method} & & C-GQA & \multicolumn{3}{c}{Method} \\ 
class & GloVe & ConceptNet & \ours\ (ours)
& & 
class & GloVe & ConceptNet & \ours\ (ours) \\ \cline{1-4} \cline{6-9}
folded book & \xmark & \cmark  & \cmark & & 
blue table & \cmark & \xmark  & \cmark   
\\
rusty truck & \xmark & \cmark  & \cmark  & & 
brown cake & \cmark & \cmark  & \xmark  
\\
small dog & \xmark & \xmark  & \cmark  & & 
balding person & \xmark & \cmark  & \cmark  
\\
eroded granite & \cmark & \cmark  & \xmark  & & 
blue tray & \xmark & \xmark  & \cmark  
\\
gray stove & \cmark & \xmark & \cmark  & & 
asian person & \xmark & \cmark  & \cmark  
\\
thick ring & \xmark & \xmark  & \xmark  & & 
yellow leaf & \xmark & \cmark  & \cmark  
\\
\bottomrule
\end{tabular}
}
\caption{Qualitative examples from the MIT-States and C-GQA datasets. A state-object pair is deemed feasible ({\cmark}) or infeasible ({\xmark}) by the respective methods.}
\label{table:appendix_qualitative}
\end{table*}

We examine qualitative examples of feasibility classifications where we compare the predictions of three methods in Table~\ref{table:appendix_qualitative}. It displays unseen classes most commonly found in the test datasets. We observe that in both the MIT-States and C-GQA datasets, our \ours\ and ConceptNet tend to predict the given classes as feasible more frequently compared to GloVe. One interesting observation arises from the class "small dog" which is surprisingly predicted as infeasible in both GloVe and ConceptNet, but correctly identified as feasible by \ours.

\end{document}